\pdfoutput=1

\documentclass[11pt]{article}
\usepackage{authblk}

\usepackage{acl}
\usepackage{graphicx}
\usepackage{algpseudocode}
\usepackage{algorithm}
\usepackage{booktabs} 
\usepackage{times}
\usepackage{latexsym}
\usepackage{hyperref}
\usepackage{pifont}
\usepackage{multirow}
\usepackage{amsmath}
\usepackage[T1]{fontenc}
\usepackage[utf8]{inputenc}
\usepackage{microtype}
\usepackage{graphicx}

\title{Improving the Calibration of Confidence Scores in Text Generation Using the Output Distribution's Characteristics}

\author{
  \textbf{Lorenzo Jaime Yu Flores}$^{1,2}$ \quad
  \textbf{Ori Ernst}$^{1,2}$ \quad
  \textbf{Jackie Chi Kit Cheung}$^{1,2,3}$ \\
  $^1$Mila - Quebec AI Institute \\
  $^2$McGill University \\
  $^3$Canada CIFAR AI Chair, Mila \\
  \texttt{\{lorenzo.flores, ori.ernst, cheungja\}@mila.quebec}
}

\begin{document}

\maketitle
\begin{abstract}
Well-calibrated model confidence scores can improve the usefulness of text generation models. For example, users can be prompted to review predictions with low confidence scores, to prevent models from returning bad or potentially dangerous predictions. However, confidence metrics are not always well calibrated in text generation. One reason is that in generation, there can be many valid answers, which previous methods do not always account for. Hence, a confident model could distribute its output probability among multiple sequences because they are all valid. We propose task-agnostic confidence metrics suited to generation, which rely solely on the probabilities associated with the model outputs without the need for further fine-tuning or heuristics. Using these, we are able to improve the calibration of BART and Flan-T5 on summarization, translation, and QA datasets. Our work can be found at \url{https://github.com/ljyflores/calibrated-confidence-for-nlg}
\end{abstract}

\section{Introduction}

Confidence scores are scores derived from a model's output, which are interpreted as the model's estimation of its own output's quality. These scores can be used in real-world applications to flag uncertain predictions in automated decision-making systems \citep{malinin2021uncertaintyestimationautoregressivestructured}, which could prompt further human review \citep{xiao2020watzeijedetecting}, or force the model to abstain from answering when unsure \citep{liu2020simpleprincipleduncertaintyestimation, kamath-etal-2020-selective}. To be useful, we want these scores to correlate with the output's quality.

A common approach to estimating confidence is through probability-based methods, which rely on the probabilities assigned by the model to output tokens. Most existing methods focus on the sequence with the highest probability, which we refer to as the top sequence \citep{murray-chiang-2018-correcting, zablotskaia-etal-2023-uncertainty, huang2023lookleapexploratorystudy, zhao-etal-2020-active, perlitz-etal-2023-active, malinin2021uncertaintyestimationautoregressivestructured}. A high probability for the top sequence suggests strong confidence in a particular prediction, while a lower value indicates uncertainty.

\begin{figure*}[ht]
\centering
  \includegraphics[width=16cm]{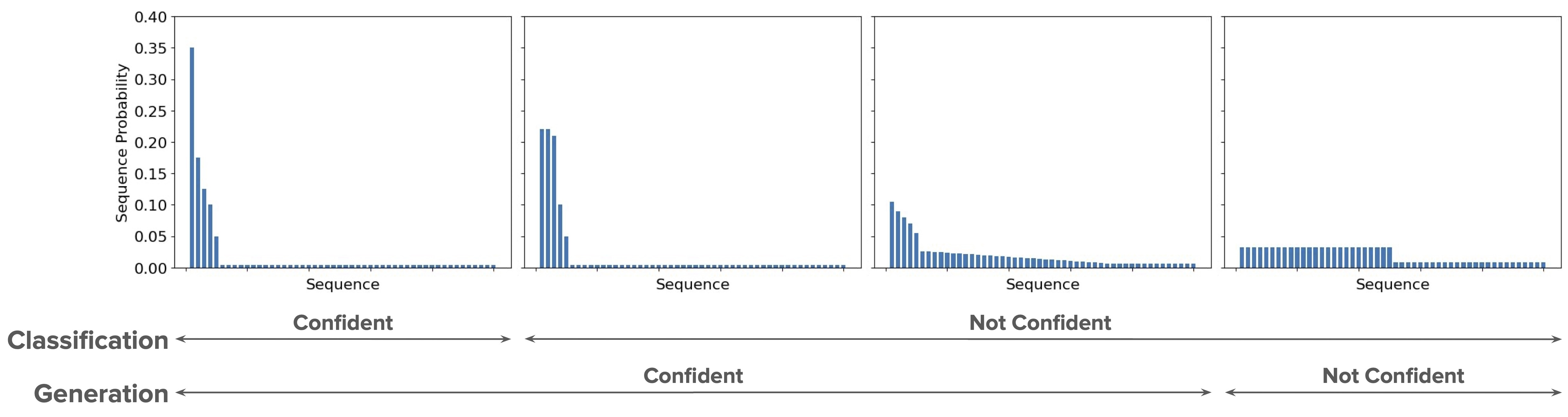}
  \caption{We illustrate the difference in interpretation of confidence in classification vs generation. Suppose a model generated output probability distributions for four different inputs; each bar is the prob. assigned to one class/sequence. In classification, only the 1st output would show model confidence, as it assigned most probability to one class. In generation, the first 3 outputs \textit{could} show confidence because multiple sequences were valid. \label{Figure:Comparison_CLS_vs_GEN}}
\end{figure*}

This approach is effective for tasks with a single correct answer. However, it faces significant challenges when applied to tasks with multiple valid outputs, as in many generation tasks. In such cases, a low top probability may not reflect a lack of confidence but rather that the model has identified several valid sequences (See Figure \ref{Figure:Comparison_CLS_vs_GEN}). Ideally, for open-ended tasks, a confident model would distribute high probabilities across multiple good sequences while assigning lower probabilities to less suitable options, while in classification, confidence can be indicated by a single high top probability.

To address this limitation, we propose new probability-based confidence estimation methods that consider the probabilities of multiple sequences instead of focusing solely on the top one. We introduce two methods: the first calculates the probability ratio between the highest-ranked sequences and the rest, while the second evaluates the thinness of the distribution's tail. Our experiments demonstrate that these metrics outperform existing baselines across three open-ended text generation tasks: translation, QA, and summarization.

\section{Related Work}

\paragraph{Probability-Based Methods} These methods rely on the model outputs to compute token-level probabilities or entropy \citep{murray-chiang-2018-correcting, zablotskaia-etal-2023-uncertainty, zhao-etal-2020-active, perlitz-etal-2023-active, Kumar2019CalibrationOE, huang2023lookleapexploratorystudy, malinin2021uncertaintyestimationautoregressivestructured}. Other work uses natural language inference models to group similar sequences before computing entropy \citep{Lin2023GeneratingWC, kuhn2023semantic, Nikitin2024KernelLE}.

\paragraph{Similarity/Disagreement Based Methods} When answers can be sampled from models (e.g., through dropout), self-consistency can be used to measure confidence: consistency across the top answers indicates confidence while variance indicates uncertainty \citep{xiao2020watzeijedetecting, Schmidt2022CombiningDG, lakshminarayanan2017simplescalablepredictiveuncertainty}.

\paragraph{Fine-Tuning Based Methods} In addition, other methods also fine-tune additional models to predict the correctness or confidence of the output \citep{Yaldiz2024DoND, kamath-etal-2020-selective, Malinin2019EnsembleDD, Fathullah2023LogitBasedED}.

\paragraph{Out of Distribution Detection (OOD) Methods} OOD can also be used to detect if a sample is in the training distribution, in which case a model is assumed to be confident  \citep{liu2020simpleprincipleduncertaintyestimation, vazhentsev-etal-2023-efficient}. 

\paragraph{Verbalized Confidence Scores} With the increased conversational ability of LLMs, recent work prompts the model to give a confidence score with its answer \citep{lin2022teaching, tian-etal-2023-just, kapoor2024largelanguagemodelstaught, han2024enhancingconfidenceexpressionlarge}.\\

Our work is closest to the probability-based methods; they are easily adaptable and task agnostic. They do not require metrics or NLI models to measure similarity, computation for OOD detection, or models that can verbalize their confidence.

\section{Method}

\paragraph{Problem Definition}

We define confidence as a score computed using the model outputs, that describes its assessment of its prediction quality. We want to compute the model's sample-level confidence for its output, that is positively correlated (i.e. calibrated) to the output's quality, measured by the evaluation metric used for the task (e.g. automated metrics, human evaluation). Formally,
\begin{align*}
    \text{Confidence}(x,\hat{y},\phi) \propto \text{Quality}(y, \hat{y}),
\end{align*}

\noindent where $x$ is the input, $y$ is the target, $\hat{y}$ is the prediction, and $\phi$ are the model parameters.

At inference time, we run beam search to generate $N$ sequences. Each sequence's probability is obtained by taking the product of the individual token probabilities. Given the i-th beam $\hat{y}^{(i)}$:

$$p_{\hat{y}^{(i)}}(x) = \prod_t p(\hat{y}_t^{(i)} | \hat{y}_{<t}^{(i)}, x)$$

\paragraph{Methods}

\begin{figure}[]
\centering
  \includegraphics[width=\columnwidth]{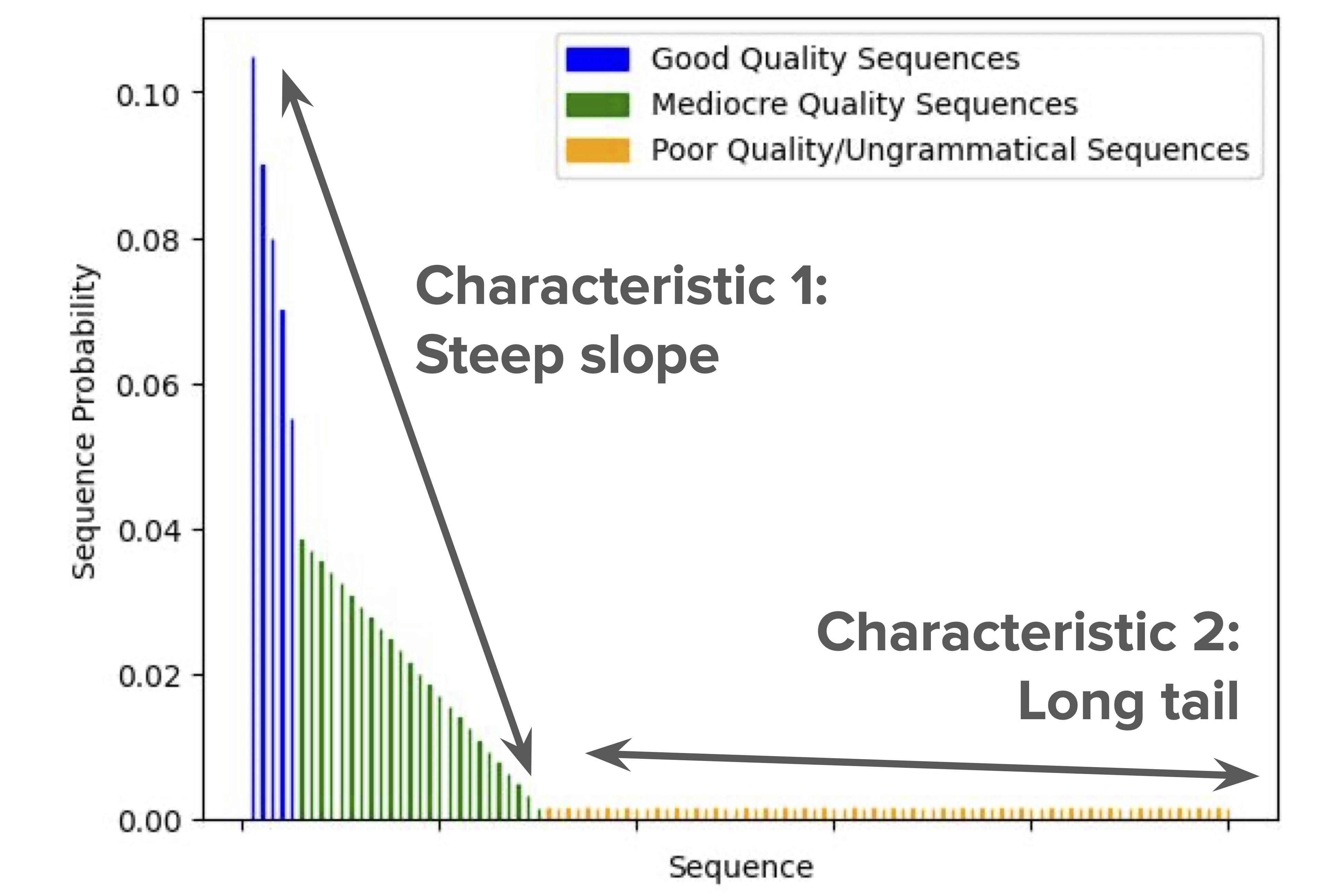}
  \caption{We hypothesize that a confident model's output \textit{would} have a steep slope and long tail; colors added for illustration purposes only. \label{Figure:Sample_Confident_Output}}
\end{figure}

We account for the fact that there can be multiple valid outputs by measuring two characteristics that we hypothesize are present in all confident outputs regardless of the number of valid sequences (See Figure~\ref{Figure:Sample_Confident_Output}). The first characteristic of a confident model is that it distinguishes good from average or bad sequences, and subsequently assigns higher probability to a select set of sequences it deems as good compared to other sequences.

\paragraph{Ratio} This motivates the ratio method: we measure how much more confident the model is in one of its \textit{best} beams $p_{\hat{y}^{(1)}}$, versus one of its \textit{average} beams $p_{\hat{y}^{(k)}}$, where $p_{\hat{y}^{(1)}}$ to $p_{\hat{y}^{(k)}}$ are sorted in descending order. This captures the intuition that a confident model will assign more probability to its best sequence than to an average sequence, whereas an unconfident model would assign similar probabilities to them. We expect the optimal value of $k$ to differ by task, as different tasks can have different levels of diversity in valid generation outputs. Hence, we tune $k$ on a validation set, and report its performance on the test set in the results.

$$\text{Ratio}(x) = \frac{p_{\hat{y}^{(1)}}(x)}{p_{\hat{y}^{(k)}}(x)}$$

\begin{figure}[]
\centering
  \includegraphics[width=\columnwidth]{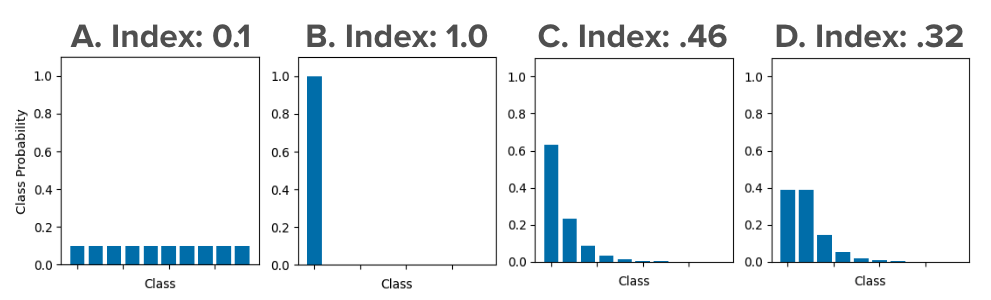}
  \caption{Samples of distributions and their tail indices \label{Figure:Tail_Index_Intuition}}
\end{figure}

The second characteristic of a confident model is that it will assign low probability to many \textit{bad} sequences. Suppose each sequence is a ``class'' – in Figure \ref{Figure:Tail_Index_Intuition}B, \ref{Figure:Tail_Index_Intuition}C, and \ref{Figure:Tail_Index_Intuition}D, the right-most classes have very small probabilities, and hence have a ``thin tail''; all three figures exhibit this property, regardless of how many \textit{correct} sequences they have. In contrast, an unconfident output where all classes receive equal probability (Figure \ref{Figure:Tail_Index_Intuition}A) has a thick tail. We quantify this with the tail index.

\paragraph{Tail Thinness} We adapt the tail index proposed by \citet{huang-tail-heaviness-2024}, originally designed to measure the thinness of statistical distributions. The higher the tail thinness, the thinner the tail. 

$$\text{Tail Thinness}(x) = \sum_{i=1}^N p_{\hat{y}^{(i)}}(x)^2$$

This sums the squared sequence probabilities for all $N$ sequences generated using beam search. Because the probabilities for $N$ sequences do not sum to 1, we first normalize them using softmax. We report the temperature used in Appendix \ref{Appendix:Implementation}. Applying this to Figure \ref{Figure:Tail_Index_Intuition}, the uniform distribution (Fig A) gets a small tail thinness, while a degenerate distribution (Fig B) has the highest tail thinness. The metric also assigns similar scores to distributions with similar tail thicknesses (Figs C and D). While this performs similarly to sequence-level entropy (Appendix \ref{Appendix:Analysis_Addtl_Baselines}), we use tail-thinness metric \cite{huang-tail-heaviness-2024} as it better describes the shape of the tail for which we have formulated our assumptions.

\section{Experiments} 

\paragraph{Fine-tuning and Inference} We first perform supervised fine-tuning (SFT) with BART Base \citep{lewis-etal-bart-2019} or Flan-T5 Base \citep{chung-etal-scale-instruction-2022}, both relatively small models with no prior ability to verbalize confidence (Appendix \ref{Appendix:Implementation}). After SFT, we generate the confidence scores for the test set. We get the sequence probabilities the top $N=100$ sequences, using beam search from HuggingFace \citep{wolf2020huggingfacestransformersstateoftheartnatural}, and replicate the baselines.

\paragraph{Evaluation}

We compute the Spearman correlation between the confidence scores and the quality scores, similar to analyses by \citet{zablotskaia-etal-2023-uncertainty, malinin2021uncertaintyestimationautoregressivestructured}. We evaluate the top beam against the reference using ROUGE-L \citep{lin-2004-rouge} for summarization, BLEU \citep{papineni-etal-2002-bleu} for translation, or F1 for question answering, and test for statistical significance with a bootstrap test \citep{berg-kirkpatrick-etal-2012-empirical}.

\paragraph{Baselines}
\label{Appendix:Baseline_Eqs}

We report the equations that we replicate from previous literature, shown in Table \ref{Table:Results_Translation}.

For the probability based methods (Rows 1-4), we compute (1) ATP: average token probability for the top sequence \citep{murray-chiang-2018-correcting, zablotskaia-etal-2023-uncertainty}, (2) ATE: average token entropy for the top sequence \citep{zhao-etal-2020-active, perlitz-etal-2023-active}, (3) DAE: dropout-based average token entropy across 10 outputs (Eq \ref{Eq:Dropout_Entropy}) \citep{malinin2021uncertaintyestimationautoregressivestructured}, and (4) WTP: weighted average of the top-K sequences' average token log probabilities (Eq \ref{Eq:Top_K}) \citep{malinin2021uncertaintyestimationautoregressivestructured}.

For the similarity/disagreement based methods (Rows 5-7), we use dropout and sample 10 outputs per instance. We compute the (1) DSM: dropout similarity using METEOR (Eq \ref{Eq:METEOR_Var}) \citep{Schmidt2022CombiningDG}, (2) DVB: dropout variance using BLEU (Eq \ref{Eq:BLEU_Var}) \citep{xiao2020watzeijedetecting}, and (3) DVK: dropout variance between token probabilities using KL divergence (Eq \ref{Eq:KL_Var}) \citep{lakshminarayanan2017simplescalablepredictiveuncertainty}.

{
    \scriptsize
    \begin{equation}
        \label{Eq:Dropout_Entropy}
        \text{Conf}_\text{DAE} = \frac{1}{10} \sum_{i=1}^{10} \frac{1}{|\hat{y}^{(i)}|} \sum_{t=1}^{|\hat{y}^{(i)}|} \mathcal{H} \left(p(\hat{y}_t^{(i)} | \hat{y}_{<t}^{(i)}, x)\right)
    \end{equation}
    
    \begin{multline*}
        \mathcal{H}\left(p(\hat{y}_t^{(i)} | \hat{y}_{<t}^{(i)}, x)\right)= \\ - \sum_{j=1}^{|\mathcal{V}|} p(\hat{y}_{t,j}^{(i)} | \hat{y}_{<t}^{(i)}, x) \text{log} \left( p(\hat{y}_{t,j}^{(i)} | \hat{y}_{<t}^{(i)}, x)\right)
    \end{multline*}
    
    \begin{equation}
        \label{Eq:Top_K}
        \text{Conf}_{\text{WTP}} = -\sum_{i=1}^{10} \pi_i \left( \frac{1}{|\hat{y}^{(i)}|} \text{ln}(p(\hat{y}^{(i)})) \right)
    \end{equation}
    
    $$\pi_i = \frac{\text{exp}\left(\frac{1}{|\hat{y}^{(i)}|} \text{ln}(p(\hat{y}^{(i)}))\right)}{\sum_{j=1}^{10} \text{exp}\left(\frac{1}{|\hat{y}^{(j)}|} \text{ln}(p(\hat{y}^{(j)}))\right)}$$
    
    $$\text{ln}(p(\hat{y}^{(i)})) = \sum_{t=1}^{|\hat{y}^{(i)}|} \text{ln}(p(\hat{y}_t^{(i)} | \hat{y}_{<t}^{(i)}, x))$$
    
    \begin{equation}
        \label{Eq:METEOR_Var}
        \text{Conf}_\text{DSM} = \frac{\sum_{i=1}^{10} \sum_{j=1}^{10} \text{Meteor}(\hat{y}^{(i)}, \hat{y}^{(j)})}{N(N-1)} 
    \end{equation}
    
    \begin{equation}
        \label{Eq:BLEU_Var}
        \text{Conf}_\text{DVB} = \sum_{i=1}^{10} \sum_{j=1}
    ^{10} (1-\text{BLEU}(\hat{y}^{(i)}, \hat{y}^{(j)}))^2
    \end{equation}
    
    \begin{equation}
        \label{Eq:KL_Var}
        \text{Conf}_\text{DVK} = \sum_{i=1}^{10} KL(p(\hat{y}^{(i)}|x), p_{\bar{y}})
    \end{equation}
    
    {
    $$\bar{y}_\text{Prob}=\frac{1}{10}\sum_{i=1}^{10} p(\hat{y}^{(i)}|x)$$
    }
}

Where $\hat{y}^{(i)}$ is the decoded sequence $i$ sampled by activating dropout, $\hat{y}_t^{(i)}$ is the $t$-th output token for sequence $i$, and $\hat{y}_{t,j}^{(i)}$ is the $j$-th vocabulary at position $t$ for sequence $i$.

\paragraph{Datasets}

We test on \textbf{Translation} (1) WMT 2017 English-German \citep{bojar-etal-2017-wmt1}, (2) WMT 2017 English-Russian \citep{bojar-etal-2017-wmt1}, (3) FLORES (Filipino Set) \citep{nllb-2022}, \textbf{Question Answering} (1) SQUAD \citep{rajpurkar-etal-2016-squad}, (2) HotpotQA \citep{yang-etal-2018-hotpotqa}, \textbf{Summarization} (1) DebateSumm \citep{roush-balaji-2020-debatesum}, (2) Reddit-TiFu \citep{kim2018abstractive}, (3) XSUM \citep{narayan-etal-2018-dont}, (4) CNN-DailyMail \citep{see-etal-2017-get}

\section{Results}

We report the correlation between the evaluation metric and confidence scores in Table \ref{Table:Results_Translation} (See Appendix \ref{Appendix:Implementation} for details). For BART, our methods achieve better correlation on 6 out of 9 datasets. We see larger gains in translation and question answering, as compared to summarization. The tail thinness method generally yields larger improvements (up to +17.2\%) than the ratio method (up to +16.1\%). For Flan-T5, our methods also achieve better correlation on 4 out of 9 datasets. Like for BART, we observe larger improvements using the tail thinness (up to +10.0\%) method than the ratio based method (up to +8.3\%). Overall, our methods yield the best performance more frequently than previous methods across all dataset-model pairs (tail thinness: 10/16, ratio: 8/16, DSM: 4/16), with median rankings of 2 and 3 for the tail and ratio methods (next being ATP, rank 4).

\begin{table*}[ht]\centering
\resizebox{\linewidth}{!}{%
    \begin{tabular}{lrrrrrrrrrrrrrrrrrrrrr}\toprule
& &\multicolumn{2}{c}{Fil–EN} &\multicolumn{2}{c}{DE–EN} &\multicolumn{2}{c}{RU–EN} &\multicolumn{2}{c}{HotpotQA} &\multicolumn{2}{c}{SQUAD} &\multicolumn{2}{c}{Debate} &\multicolumn{2}{c}{Reddit} &\multicolumn{2}{c}{CNN} & \multicolumn{2}{c}{XSUM} & \multicolumn{2}{c}{Rank} \\\cmidrule{3-22}
& &Bt &FT5 &Bt &FT5 &Bt &FT5 &Bt &FT5 &Bt &FT5 &Bt &FT5 &Bt &FT5 &Bt &FT5 &Bt &FT5 & Avg & Med \\\midrule
\parbox[t]{4mm}{\multirow{4}{*}{\rotatebox[origin=c]{90}{Probability}}} &ATP &.473 &.468 &.028 &.370 &.530 &.023 &.209 &.302 &.391 & \textcolor{blue}{.577} &.447 &.247 &\textbf{.618} &\textbf{.577} &.109 &.156 &.119 &.078  & 4.4 & 4 \\
&ATE &.308 &.335 &.035 &.297 &.437 &.042 &.051 &.152 &.094 &.049 &.416 &.248 &.615 &.474 &.020 &.138 &.093 &.082 & 6.4 & 7 \\
&DAE &.217 &.161 &.346 &.294 &.230 &.178 & \textcolor{blue}{.242} & \textcolor{blue}{.367} &.327 &.226 &.135 &.037 &.049 &.049 &\textbf{.295} &\textbf{.380} &.314 &.353 & 5.4 & 6 \\
&WTP & \textcolor{blue}{.516} &.495 &.162 &.287 & \textcolor{blue}{.602} &.055 &.130 &.180 &.179 &.020 & \textcolor{blue}{.489} & \textcolor{blue}{.253} &.616 &.575 &.106 &.162 &.120 &.063 & 5 & 5 \\
\midrule
\parbox[t]{4mm}{\multirow{3}{*}{\rotatebox[origin=c]{90}{Sim/Diff}}} &DSM &.441 &\textbf{.508} &.424 &\textbf{.462} &.374 &.486 &.168 &.270 & \textcolor{blue}{.394} &.332 &.192 &.038 &.038 &.167 &.255 &.323 &\textbf{.323} &\textbf{.383} & 4.4 & 4.5 \\
&DVB &.455 &.489 & \textcolor{blue}{.512} &.461 &.409 & \textcolor{blue}{.488} &.043 &.000 &.378 &.467 &.144 &.061 &.058 &.143 &.264 &.325 &.305 &.363 & 4.7 & 4.5 \\
&DVK &.001 &.008 &.110 &.064 &.110 &.013 &.177 &.232 &.340 &.426 &.063 &.025 &.045 &.059 &.065 &.070 &.103 &.117 & 7.6 & 8 \\
\midrule
\parbox[t]{4mm}{\multirow{2}{*}{\rotatebox[origin=c]{90}{Ours}}} &Ratio &\textbf{.546} &.200 & \textsuperscript{\ding{72}}\textbf{.653} &.209 & \textsuperscript{\ding{72}}\textbf{.768} &\textbf{.491} &\textbf{.249} &.360 & \textsuperscript{\ding{72}}\textbf{.505} &.565 &\textbf{.496} & \textsuperscript{\ding{73}}\textbf{.293} &.596 &.304 &.103 &.055 &.082 &.196 & 3.9 & 3 \\
&Tail & \textsuperscript{\ding{72}}\textbf{.649} &.380 & \textsuperscript{\ding{72}}\textbf{.648} &.190 & \textsuperscript{\ding{72}}\textbf{.779} &\textbf{.506} &\textbf{.255} & \textsuperscript{\ding{72}}\textbf{.451} & \textsuperscript{\ding{72}}\textbf{.493} &\textbf{.582} &\textbf{.518} & \textbf{\textsuperscript{\ding{72}}.354} &.601 &.300 &.100 &.031 &.131 &.212 & 3.2 & 2 \\
\bottomrule
\end{tabular}
}
\caption{Spearman correlation (absolute value) between confidence score and evaluation metric/quality score (BLEU for translation, F1 for QA, RougeL for summarization); Bt: BART, FT5: Flan-T5, stars indicate significant difference from \textcolor{blue}{next best method} (bootstrap test, \textsuperscript{\ding{73}}$\alpha=0.10$, \textsuperscript{\ding{72}}$\alpha=0.05$)}
\label{Table:Results_Translation}
\end{table*}

\begin{figure*}[]
\centering
  \includegraphics[width=\linewidth]{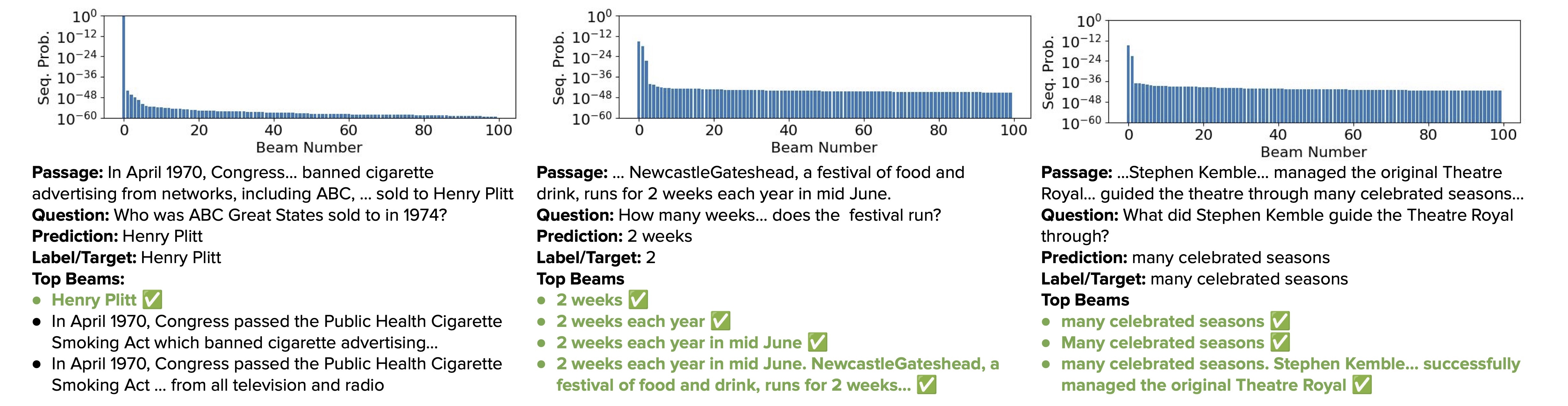}
  \caption{Samples from SQUAD \citep{rajpurkar-etal-2016-squad}; 1st image only has one valid output, whereas the 2nd and 3rd have multiple; our tail-thinness and ratio based confidence correctly assign high confidence to all samples, but avg. log prob. only assigns high confidence to the first image (Note: The y-axis is plotted on the log scale) \label{Figure:Multiple_Outputs}}
\end{figure*}

\paragraph{Robustness to Multiple Valid Sequences} Qualitatively, we find that our methods assign high confidence to outputs where there are multiple valid sequences. We look at examples where our metrics assigned high confidence, but other methods like average token log probability assigned low confidence (See Figure \ref{Figure:Multiple_Outputs}). In these examples, there were indeed multiple, correct outputs; this resulted in lower probability for the top beam (2nd and 3rd image). If we only used the top beam's probability to measure confidence, we might conclude that the model is unconfident. In contrast, our methods which rely on the ratio of sequence probabilities and tail thinness, rather than the top probability, are able to correctly identify that the model is still confident in such scenarios. This illustrates how using features of the distribution like slope or tail thinness can be more indicative of confidence in text generation, rather than solely looking at the features of the top output.

\paragraph{Failure Cases} We examine samples for which the confidence scores are not well calibrated. Looking at the FLORES (Filipino) for Flan-T5, we observed samples where the model was confident, but its output was bad. Here, the model failed to translate a few key terms, which changed the meaning of the sentence (See Table \ref{Table:Miscalibrated_Model}). Other times, the confidence scores were well calibrated, but the quality score was not estimated well. This stemmed from noisy labels or limitations of the evaluation metric (See Table \ref{Table:Miscalibrated_Label}) which may require future work.

\begin{figure*}[!htb]
\centering
  \includegraphics[width=15cm]{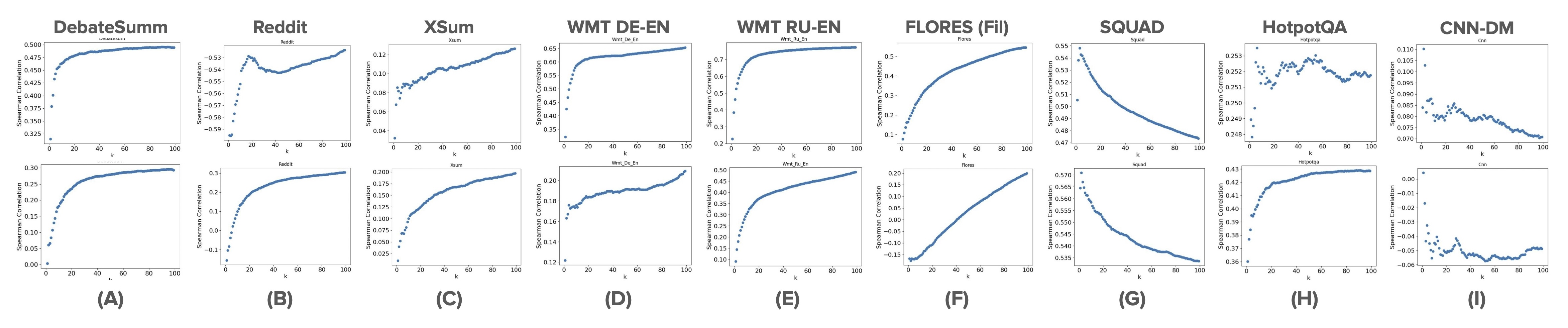}
  \caption{Spearman Correlation vs $k$ on test set for BART (top row) and Flan-T5 (bottom row); In general, open-ended tasks (summarization: A-C, translation: D-F) benefit from larger $k$, close-ended tasks (QA: G-H, Reddit: I) use smaller $k$ \label{Figure:Correlation_by_k}}
\end{figure*}

\paragraph{Choice of $k$} In general, open ended tasks (translation, summarization) benefited from larger values for $k$, and close-ended tasks (QA) from smaller values of $k$ (See Figure \ref{Figure:Correlation_by_k}). One explanation for this could be that $k$ serves as a parameter which delineates the \textit{good} vs. \textit{average} sequences. Finding the $k$ that best separates the two groups allows us to most accurately measure the difference in confidence between both groups. Open-ended tasks can have more good sequences, hence correlation is maximized when we choose a higher value for $k$. In contrast, close-ended tasks have fewer good sequences, so a lower value for $k$ is better. 

We note that the optimal value for $k$ may be very large for open-ended tasks like summarization. In our experiments, we capped $k$ to 100 due to computational limitations, which may have underestimated its optimal value, and thus explain why our methods are less competitive than the baselines.

\section{Conclusion}

We identified characteristics of output distributions from a confident model in generation tasks, and used this to propose metrics that capture these characteristics. We find that on various datasets, these characteristics are better correlated to quality metrics than previous methods. 

There are various directions future work can take to improve the computation and evaluation of confidence scores. For example, we observed that models can be overly confident in wrong answers (Table \ref{Table:Miscalibrated_Model}), which can lead to miscalibration when using our methods. Future work can study the reasons for model overconfidence, and propose ways to reduce or account for this. In addition, more work is required to improve the evaluation of calibration in generation tasks. Concretely, calibration is well defined for classification tasks when there is a binary outcome, and metrics like expected calibration error \cite{tian-etal-2023-just} measure whether the model's confidence scores match its accuracy (e.g. A model should be right in 90\% of the examples for which it claimed to be 90\% confident). More work can be done to develop metrics that capture a similar notion of calibration in generation tasks where the outcome is continuous.

\section*{Limitations and Potential Risks} 
One limitation is that our methods require users to compute $k$ beams, both when tuning hyperparameters, and at inference time. This can be computationally expensive, especially for open-ended tasks where the optimal value of $k$ may be large. Hence, future work may find more efficient ways of estimating the confidence.

Another caveat is that we only evaluate the performance of our methods on a limited set of models and parameters. For example, we fine-tuned various models with early stopping. Hence, future work which seeks to use these methods must re-evaluate the scores on their tasks, to avoid deploying miscalibrated confidence scores in practical settings.

Finally, future work could study better ways to evaluate confidence scores; we found that traditional evaluation metrics may lead to poor quality ratings, and it was difficult to find datasets with human evaluation scores to use.

\section*{Acknowledgements}

We would like to thank Ines Arous, Ziling Cheng, Zichao Li, and Caleb Moses for providing useful writing feedback, and David Austin, Cesare Spinoso-di Piano, and Xiyuan Zou for engaging in productive discussions as we developed the paper. We also thank NVIDIA for their support in terms of computational resources. This work was also supported in part by the IVADO Postdoctoral Fellowship.

\bibliography{anthology,custom}

\begin{thebibliography}{38}
\expandafter\ifx\csname natexlab\endcsname\relax\def\natexlab#1{#1}\fi

\bibitem[{Berg-Kirkpatrick et~al.(2012)Berg-Kirkpatrick, Burkett, and Klein}]{berg-kirkpatrick-etal-2012-empirical}
Taylor Berg-Kirkpatrick, David Burkett, and Dan Klein. 2012.
\newblock \href {https://aclanthology.org/D12-1091/} {An empirical investigation of statistical significance in {NLP}}.
\newblock In \emph{Proceedings of the 2012 Joint Conference on Empirical Methods in Natural Language Processing and Computational Natural Language Learning}, pages 995--1005, Jeju Island, Korea. Association for Computational Linguistics.

\bibitem[{Bojar et~al.(2017)Bojar, Chatterjee, Federmann, Graham, Haddow, Huang, Huck, Koehn, Liu, Logacheva, Monz, Negri, Post, Rubino, Specia, and Turchi}]{bojar-etal-2017-wmt1}
Ond~{r}ej Bojar, Rajen Chatterjee, Christian Federmann, Yvette Graham, Barry Haddow, Shujian Huang, Matthias Huck, Philipp Koehn, Qun Liu, Varvara Logacheva, Christof Monz, Matteo Negri, Matt Post, Raphael Rubino, Lucia Specia, and Marco Turchi. 2017.
\newblock \href {http://www.aclweb.org/anthology/W17-4717} {Findings of the 2017 conference on machine translation (wmt17)}.
\newblock In \emph{Proceedings of the Second Conference on Machine Translation, Volume 2: Shared Task Papers}, pages 169--214, Copenhagen, Denmark. Association for Computational Linguistics.

\bibitem[{Chung et~al.(2022)Chung, Hou, Longpre, Zoph, Tay, Fedus, Li, Wang, Dehghani, Brahma, Webson, Gu, Dai, Suzgun, Chen, Chowdhery, Narang, Mishra, Yu, Zhao, Huang, Dai, Yu, Petrov, Chi, Dean, Devlin, Roberts, Zhou, Le, and Wei}]{chung-etal-scale-instruction-2022}
Hyung~Won Chung, Le~Hou, Shayne Longpre, Barret Zoph, Yi~Tay, William Fedus, Eric Li, Xuezhi Wang, Mostafa Dehghani, Siddhartha Brahma, Albert Webson, Shixiang~Shane Gu, Zhuyun Dai, Mirac Suzgun, Xinyun Chen, Aakanksha Chowdhery, Sharan Narang, Gaurav Mishra, Adams Yu, Vincent Zhao, Yanping Huang, Andrew Dai, Hongkun Yu, Slav Petrov, Ed~H. Chi, Jeff Dean, Jacob Devlin, Adam Roberts, Denny Zhou, Quoc~V. Le, and Jason Wei. 2022.
\newblock \href {https://doi.org/10.48550/ARXIV.2210.11416} {Scaling instruction-finetuned language models}.

\bibitem[{Fathullah et~al.(2023)Fathullah, Xia, and Gales}]{Fathullah2023LogitBasedED}
Yassir Fathullah, Guoxuan Xia, and Mark John~Francis Gales. 2023.
\newblock \href {https://api.semanticscholar.org/CorpusID:258741024} {Logit-based ensemble distribution distillation for robust autoregressive sequence uncertainties}.
\newblock \emph{ArXiv}, abs/2305.10384.

\bibitem[{Han et~al.(2024)Han, Li, Chen, Shi, Du, Xiao, Liang, and Lin}]{han2024enhancingconfidenceexpressionlarge}
Haixia Han, Tingyun Li, Shisong Chen, Jie Shi, Chengyu Du, Yanghua Xiao, Jiaqing Liang, and Xin Lin. 2024.
\newblock \href {http://arxiv.org/abs/2404.10315} {Enhancing confidence expression in large language models through learning from past experience}.

\bibitem[{Huang(2024)}]{huang-tail-heaviness-2024}
Hening Huang. 2024.
\newblock \href {https://doi.org/10.32388/9B8HK9} {A new measure of the tail-heaviness of a probability distribution}.

\bibitem[{Huang et~al.(2023)Huang, Song, Wang, Zhao, Chen, Juefei-Xu, and Ma}]{huang2023lookleapexploratorystudy}
Yuheng Huang, Jiayang Song, Zhijie Wang, Shengming Zhao, Huaming Chen, Felix Juefei-Xu, and Lei Ma. 2023.
\newblock \href {http://arxiv.org/abs/2307.10236} {Look before you leap: An exploratory study of uncertainty measurement for large language models}.

\bibitem[{Kamath et~al.(2020)Kamath, Jia, and Liang}]{kamath-etal-2020-selective}
Amita Kamath, Robin Jia, and Percy Liang. 2020.
\newblock \href {https://doi.org/10.18653/v1/2020.acl-main.503} {Selective question answering under domain shift}.
\newblock In \emph{Proceedings of the 58th Annual Meeting of the Association for Computational Linguistics}, pages 5684--5696, Online. Association for Computational Linguistics.

\bibitem[{Kapoor et~al.(2024)Kapoor, Gruver, Roberts, Collins, Pal, Bhatt, Weller, Dooley, Goldblum, and Wilson}]{kapoor2024largelanguagemodelstaught}
Sanyam Kapoor, Nate Gruver, Manley Roberts, Katherine Collins, Arka Pal, Umang Bhatt, Adrian Weller, Samuel Dooley, Micah Goldblum, and Andrew~Gordon Wilson. 2024.
\newblock \href {http://arxiv.org/abs/2406.08391} {Large language models must be taught to know what they don't know}.

\bibitem[{Kim et~al.(2018)Kim, Kim, and Kim}]{kim2018abstractive}
Byeongchang Kim, Hyunwoo Kim, and Gunhee Kim. 2018.
\newblock \href {http://arxiv.org/abs/1811.00783} {Abstractive summarization of reddit posts with multi-level memory networks}.

\bibitem[{Kuhn et~al.(2023)Kuhn, Gal, and Farquhar}]{kuhn2023semantic}
Lorenz Kuhn, Yarin Gal, and Sebastian Farquhar. 2023.
\newblock \href {https://openreview.net/forum?id=VD-AYtP0dve} {Semantic uncertainty: Linguistic invariances for uncertainty estimation in natural language generation}.
\newblock In \emph{The Eleventh International Conference on Learning Representations}.

\bibitem[{Kumar and Sarawagi(2019)}]{Kumar2019CalibrationOE}
Aviral Kumar and Sunita Sarawagi. 2019.
\newblock \href {https://api.semanticscholar.org/CorpusID:67855916} {Calibration of encoder decoder models for neural machine translation}.
\newblock \emph{ArXiv}, abs/1903.00802.

\bibitem[{Lakshminarayanan et~al.(2017)Lakshminarayanan, Pritzel, and Blundell}]{lakshminarayanan2017simplescalablepredictiveuncertainty}
Balaji Lakshminarayanan, Alexander Pritzel, and Charles Blundell. 2017.
\newblock \href {http://arxiv.org/abs/1612.01474} {Simple and scalable predictive uncertainty estimation using deep ensembles}.

\bibitem[{Lewis et~al.(2019)Lewis, Liu, Goyal, Ghazvininejad, Mohamed, Levy, Stoyanov, and Zettlemoyer}]{lewis-etal-bart-2019}
Mike Lewis, Yinhan Liu, Naman Goyal, Marjan Ghazvininejad, Abdelrahman Mohamed, Omer Levy, Veselin Stoyanov, and Luke Zettlemoyer. 2019.
\newblock \href {http://arxiv.org/abs/1910.13461} {{BART:} denoising sequence-to-sequence pre-training for natural language generation, translation, and comprehension}.
\newblock \emph{CoRR}, abs/1910.13461.

\bibitem[{Lin(2004)}]{lin-2004-rouge}
Chin-Yew Lin. 2004.
\newblock \href {https://aclanthology.org/W04-1013/} {{ROUGE}: A package for automatic evaluation of summaries}.
\newblock In \emph{Text Summarization Branches Out}, pages 74--81, Barcelona, Spain. Association for Computational Linguistics.

\bibitem[{Lin et~al.(2022)Lin, Hilton, and Evans}]{lin2022teaching}
Stephanie Lin, Jacob Hilton, and Owain Evans. 2022.
\newblock \href {https://openreview.net/forum?id=8s8K2UZGTZ} {Teaching models to express their uncertainty in words}.
\newblock \emph{Transactions on Machine Learning Research}.

\bibitem[{Lin et~al.(2023)Lin, Trivedi, and Sun}]{Lin2023GeneratingWC}
Zhen Lin, Shubhendu Trivedi, and Jimeng Sun. 2023.
\newblock \href {https://api.semanticscholar.org/CorpusID:258967487} {Generating with confidence: Uncertainty quantification for black-box large language models}.
\newblock \emph{Trans. Mach. Learn. Res.}, 2024.

\bibitem[{Liu et~al.(2020)Liu, Lin, Padhy, Tran, Bedrax-Weiss, and Lakshminarayanan}]{liu2020simpleprincipleduncertaintyestimation}
Jeremiah~Zhe Liu, Zi~Lin, Shreyas Padhy, Dustin Tran, Tania Bedrax-Weiss, and Balaji Lakshminarayanan. 2020.
\newblock \href {http://arxiv.org/abs/2006.10108} {Simple and principled uncertainty estimation with deterministic deep learning via distance awareness}.

\bibitem[{Malinin and Gales(2021)}]{malinin2021uncertaintyestimationautoregressivestructured}
Andrey Malinin and Mark Gales. 2021.
\newblock \href {http://arxiv.org/abs/2002.07650} {Uncertainty estimation in autoregressive structured prediction}.

\bibitem[{Malinin et~al.(2019)Malinin, Mlodozeniec, and Gales}]{Malinin2019EnsembleDD}
Andrey Malinin, Bruno Mlodozeniec, and Mark John~Francis Gales. 2019.
\newblock \href {https://api.semanticscholar.org/CorpusID:141465546} {Ensemble distribution distillation}.
\newblock \emph{ArXiv}, abs/1905.00076.

\bibitem[{Murray and Chiang(2018)}]{murray-chiang-2018-correcting}
Kenton Murray and David Chiang. 2018.
\newblock \href {https://doi.org/10.18653/v1/W18-6322} {Correcting length bias in neural machine translation}.
\newblock In \emph{Proceedings of the Third Conference on Machine Translation: Research Papers}, pages 212--223, Brussels, Belgium. Association for Computational Linguistics.

\bibitem[{Narayan et~al.(2018)Narayan, Cohen, and Lapata}]{narayan-etal-2018-dont}
Shashi Narayan, Shay~B. Cohen, and Mirella Lapata. 2018.
\newblock \href {https://doi.org/10.18653/v1/D18-1206} {Don`t give me the details, just the summary! topic-aware convolutional neural networks for extreme summarization}.
\newblock In \emph{Proceedings of the 2018 Conference on Empirical Methods in Natural Language Processing}, pages 1797--1807, Brussels, Belgium. Association for Computational Linguistics.

\bibitem[{Nikitin et~al.(2024)Nikitin, Kossen, Gal, and Marttinen}]{Nikitin2024KernelLE}
Alexander Nikitin, Jannik Kossen, Yarin Gal, and Pekka Marttinen. 2024.
\newblock \href {https://api.semanticscholar.org/CorpusID:270123445} {Kernel language entropy: Fine-grained uncertainty quantification for llms from semantic similarities}.
\newblock \emph{ArXiv}, abs/2405.20003.

\bibitem[{NLLB(2022)}]{nllb-2022}
Team NLLB. 2022.
\newblock No language left behind: Scaling human-centered machine translation.

\bibitem[{Papineni et~al.(2002)Papineni, Roukos, Ward, and Zhu}]{papineni-etal-2002-bleu}
Kishore Papineni, Salim Roukos, Todd Ward, and Wei-Jing Zhu. 2002.
\newblock \href {https://doi.org/10.3115/1073083.1073135} {{B}leu: a method for automatic evaluation of machine translation}.
\newblock In \emph{Proceedings of the 40th Annual Meeting of the Association for Computational Linguistics}, pages 311--318, Philadelphia, Pennsylvania, USA. Association for Computational Linguistics.

\bibitem[{Perlitz et~al.(2023)Perlitz, Gera, Shmueli-Scheuer, Sheinwald, Slonim, and Ein-Dor}]{perlitz-etal-2023-active}
Yotam Perlitz, Ariel Gera, Michal Shmueli-Scheuer, Dafna Sheinwald, Noam Slonim, and Liat Ein-Dor. 2023.
\newblock \href {https://doi.org/10.18653/v1/2023.emnlp-main.611} {Active learning for natural language generation}.
\newblock In \emph{Proceedings of the 2023 Conference on Empirical Methods in Natural Language Processing}, pages 9862--9877, Singapore. Association for Computational Linguistics.

\bibitem[{Rajpurkar et~al.(2016)Rajpurkar, Zhang, Lopyrev, and Liang}]{rajpurkar-etal-2016-squad}
Pranav Rajpurkar, Jian Zhang, Konstantin Lopyrev, and Percy Liang. 2016.
\newblock \href {https://doi.org/10.18653/v1/D16-1264} {{SQ}u{AD}: 100,000+ questions for machine comprehension of text}.
\newblock In \emph{Proceedings of the 2016 Conference on Empirical Methods in Natural Language Processing}, pages 2383--2392, Austin, Texas. Association for Computational Linguistics.

\bibitem[{Roush and Balaji(2020)}]{roush-balaji-2020-debatesum}
Allen Roush and Arvind Balaji. 2020.
\newblock \href {https://aclanthology.org/2020.argmining-1.1/} {{D}ebate{S}um: A large-scale argument mining and summarization dataset}.
\newblock In \emph{Proceedings of the 7th Workshop on Argument Mining}, pages 1--7, Online. Association for Computational Linguistics.

\bibitem[{Schmidt et~al.(2022)Schmidt, Bartezzaghi, Bogojeska, Malossi, and Vu}]{Schmidt2022CombiningDG}
Maximilian Schmidt, A.~Bartezzaghi, Jasmina Bogojeska, Adelmo Cristiano~Innocenza Malossi, and Thang Vu. 2022.
\newblock \href {https://api.semanticscholar.org/CorpusID:254044648} {Combining data generation and active learning for low-resource question answering}.
\newblock In \emph{International Conference on Artificial Neural Networks}.

\bibitem[{See et~al.(2017)See, Liu, and Manning}]{see-etal-2017-get}
Abigail See, Peter~J. Liu, and Christopher~D. Manning. 2017.
\newblock \href {https://doi.org/10.18653/v1/P17-1099} {Get to the point: Summarization with pointer-generator networks}.
\newblock In \emph{Proceedings of the 55th Annual Meeting of the Association for Computational Linguistics (Volume 1: Long Papers)}, pages 1073--1083, Vancouver, Canada. Association for Computational Linguistics.

\bibitem[{Tian et~al.(2023)Tian, Mitchell, Zhou, Sharma, Rafailov, Yao, Finn, and Manning}]{tian-etal-2023-just}
Katherine Tian, Eric Mitchell, Allan Zhou, Archit Sharma, Rafael Rafailov, Huaxiu Yao, Chelsea Finn, and Christopher Manning. 2023.
\newblock \href {https://doi.org/10.18653/v1/2023.emnlp-main.330} {Just ask for calibration: Strategies for eliciting calibrated confidence scores from language models fine-tuned with human feedback}.
\newblock In \emph{Proceedings of the 2023 Conference on Empirical Methods in Natural Language Processing}, pages 5433--5442, Singapore. Association for Computational Linguistics.

\bibitem[{Vazhentsev et~al.(2023)Vazhentsev, Tsvigun, Vashurin, Petrakov, Vasilev, Panov, Panchenko, and Shelmanov}]{vazhentsev-etal-2023-efficient}
Artem Vazhentsev, Akim Tsvigun, Roman Vashurin, Sergey Petrakov, Daniil Vasilev, Maxim Panov, Alexander Panchenko, and Artem Shelmanov. 2023.
\newblock \href {https://doi.org/10.18653/v1/2023.findings-acl.93} {Efficient out-of-domain detection for sequence to sequence models}.
\newblock In \emph{Findings of the Association for Computational Linguistics: ACL 2023}, pages 1430--1454, Toronto, Canada. Association for Computational Linguistics.

\bibitem[{Wolf et~al.(2020)Wolf, Debut, Sanh, Chaumond, Delangue, Moi, Cistac, Rault, Louf, Funtowicz, Davison, Shleifer, von Platen, Ma, Jernite, Plu, Xu, Scao, Gugger, Drame, Lhoest, and Rush}]{wolf2020huggingfacestransformersstateoftheartnatural}
Thomas Wolf, Lysandre Debut, Victor Sanh, Julien Chaumond, Clement Delangue, Anthony Moi, Pierric Cistac, Tim Rault, Rémi Louf, Morgan Funtowicz, Joe Davison, Sam Shleifer, Patrick von Platen, Clara Ma, Yacine Jernite, Julien Plu, Canwen Xu, Teven~Le Scao, Sylvain Gugger, Mariama Drame, Quentin Lhoest, and Alexander~M. Rush. 2020.
\newblock \href {http://arxiv.org/abs/1910.03771} {Huggingface's transformers: State-of-the-art natural language processing}.

\bibitem[{Xiao et~al.(2020)Xiao, Gomez, and Gal}]{xiao2020watzeijedetecting}
Tim~Z. Xiao, Aidan~N. Gomez, and Yarin Gal. 2020.
\newblock \href {http://arxiv.org/abs/2006.08344} {Wat zei je? detecting out-of-distribution translations with variational transformers}.

\bibitem[{Yaldiz et~al.(2024)Yaldiz, Bakman, Buyukates, Tao, Ramakrishna, Dimitriadis, and Avestimehr}]{Yaldiz2024DoND}
Duygu~Nur Yaldiz, Yavuz~Faruk Bakman, Baturalp Buyukates, Chenyang Tao, Anil Ramakrishna, Dimitrios Dimitriadis, and Amir~Salman Avestimehr. 2024.
\newblock \href {https://api.semanticscholar.org/CorpusID:270560969} {Do not design, learn: A trainable scoring function for uncertainty estimation in generative llms}.
\newblock \emph{ArXiv}, abs/2406.11278.

\bibitem[{Yang et~al.(2018)Yang, Qi, Zhang, Bengio, Cohen, Salakhutdinov, and Manning}]{yang-etal-2018-hotpotqa}
Zhilin Yang, Peng Qi, Saizheng Zhang, Yoshua Bengio, William Cohen, Ruslan Salakhutdinov, and Christopher~D. Manning. 2018.
\newblock \href {https://doi.org/10.18653/v1/D18-1259} {{H}otpot{QA}: A dataset for diverse, explainable multi-hop question answering}.
\newblock In \emph{Proceedings of the 2018 Conference on Empirical Methods in Natural Language Processing}, pages 2369--2380, Brussels, Belgium. Association for Computational Linguistics.

\bibitem[{Zablotskaia et~al.(2023)Zablotskaia, Phan, Maynez, Narayan, Ren, and Liu}]{zablotskaia-etal-2023-uncertainty}
Polina Zablotskaia, Du~Phan, Joshua Maynez, Shashi Narayan, Jie Ren, and Jeremiah Liu. 2023.
\newblock \href {https://doi.org/10.18653/v1/2023.findings-emnlp.197} {On uncertainty calibration and selective generation in probabilistic neural summarization: A benchmark study}.
\newblock In \emph{Findings of the Association for Computational Linguistics: EMNLP 2023}, pages 2980--2992, Singapore. Association for Computational Linguistics.

\bibitem[{Zhao et~al.(2020)Zhao, Zhang, Zhou, and Zhang}]{zhao-etal-2020-active}
Yuekai Zhao, Haoran Zhang, Shuchang Zhou, and Zhihua Zhang. 2020.
\newblock \href {https://doi.org/10.18653/v1/2020.findings-emnlp.162} {Active learning approaches to enhancing neural machine translation}.
\newblock In \emph{Findings of the Association for Computational Linguistics: EMNLP 2020}, pages 1796--1806, Online. Association for Computational Linguistics.

\end{thebibliography}
\bibliographystyle{acl_natbib}

\appendix

\section{Implementation Details}
\label{Appendix:Implementation}

All models were fine-tuned on one NVIDIA A100 GPU, with a constant learning rate 5e-5, and batch size of 10. The scripts and fine-tuned models are provided in the repository. Roughly 80 hours were used to train and perform inference on one GPU.

During SFT, we train for at most 3 epochs. We observe overfitting on many datasets, and remedy this by employing early stopping, where we stop training if the loss on the validation set does not improve after 2 steps. This was applied to all datasets except HotpotQA, WMT RU-EN, and DebateSumm. We report the number of SFT steps in Table \ref{Table:FT_Params}.

\begin{table}[ht]\centering
\begin{tabular}{lll}\toprule
Dataset & BART & Flan-T5 \\
\midrule
WMT DE-EN & 200 & 200 \\
WMT RU-EN & 6000 & 6000 \\
FLORES Filipino & 260 & 200 \\
\midrule
SQUAD & 220 & 240 \\
HotpotQA & 26835 & 26835 \\
\midrule
DebateSumm & 1500 & 1500 \\
Reddit & 140 & 200 \\
CNN & 200 & 200 \\
XSUM & 120 & 200 \\
\bottomrule
\end{tabular}
\caption{Number of Fine-Tuning Steps Taken per Task and Model}
\label{Table:FT_Params}
\end{table}

We report the parameters used for the ratio and tail-thinness methods ($k$: ratio method, temperature: softmax for the tail method) in Table \ref{Table:Parameters}.

\begin{table}[!htb]\centering
\begin{tabular}{llll}\toprule
Dataset & Model & $k$ & Temp \\
\midrule
FLORES Filipino & BART      & 99  & 1.000 \\
                & Flan-T5   & 99  & 1.000 \\
WMT DE-EN       & BART      & 99  & 1.000 \\
                & Flan-T5   & 99  & 1.000 \\
WMT RU-EN       & BART      & 79  & 1.000 \\
                & Flan-T5   & 99  & 1.000 \\
\midrule
HotpotQA        & BART      & 1  & 0.010 \\
                & Flan-T5   & 1  & 0.050 \\
SQUAD           & BART      & 1   & 0.050 \\
                & Flan-T5   & 4   & 0.001 \\
\midrule
DebateSumm      & BART      & 95 & 1.000 \\
                & Flan-T5   & 85 & 1.000 \\
Reddit          & BART      & 2  & 0.005 \\
                & Flan-T5   & 99  & 0.010 \\
CNN             & BART      & 3   & 0.001 \\
                & Flan-T5   & 77  & 0.001 \\
XSUM            & BART      & 4   & 0.100 \\
                & Flan-T5   & 98  & 0.100 \\
\bottomrule
\end{tabular}
\caption{Fine-Tuning Parameters for Various Tasks}
\label{Table:Parameters}
\end{table}

\section{Dataset Details}

\paragraph{Licenses} The FLORES, SQUAD, and HotpotQA datasets were used under the Creative Commons Attribution Share Alike 4.0 license; DebateSumm, XSUM, and Reddit-TiFu used the MIT license, the CNN DailyMail dataset used Apache2.0, and WMT17 did not provide a license on the HuggingFace platform.

\paragraph{Data Splits} For training and inference efficiency, we only use subsets of the datasets in some cases. The scripts used to generate the datasets are provided in the repository. At a high level, we take and shuffle the original dataset, then generate a train and test split from that. We perform inference on the test set, for which we report the statistics in the results section. Note that because we employ early stopping, the full training set is not necessarily provided. The number of steps actually taken are reported in Appendix \ref{Appendix:Implementation}.

\begin{table}[!htb]\centering
\begin{tabular}{llll}\toprule
Dataset & Train & Val & Test \\
\midrule
FLORES Filipino & 900   & 97  & 1012 \\
WMT DE-EN       & 2000  & 100 & 1000\\
WMT RU-EN       & 20000 & 100 & 1000 \\
\midrule
HotpotQA        & 89447 & 100 & 1000 \\
SQUAD           & 87599 & 100 & 1000 \\
\midrule
DebateSumm      & 5000  & 100 & 1000 \\
Reddit          & 2000  & 100 & 1000 \\
CNN             & 2000  & 100 & 1000 \\
XSUM            & 20000 & 100 & 1000 \\
\bottomrule
\end{tabular}
\caption{Data Splits by Task}
\label{Table:Data_Splits}
\end{table}

\section{Analysis of Additional Baselines}
\label{Appendix:Analysis_Addtl_Baselines}
In addition to the baselines from previous literature, we compare our method to two baselines that can be viewed as straightforward methods of creating a confidence score, namely beam-level entropy (Eq \ref{Eq:beam_entropy}) and the sum of the top-$k$ beam probabilities (Eq \ref{Eq:top_k_probs}), and report the results in Table \ref{Table:Additional_Baselines}.

Overall, we find that the performance of the tail method is similar to that of beam-level entropy, although the original tail method performs slightly better in multiple cases.

As for the Top-K baseline, both our ratio and tail methods outperform it on translation and QA tasks for BART, and on QA, English-Russian, DebateSumm, and XSum for T5. However, it underperforms top-K on other tasks, particularly in summarization, which is similar to our findings for the WTP method presented in the main paper.

{
    \scriptsize
    $$p(\hat{y}^{(i)}) = \sum_{t=1}^{|\hat{y}^{(i)}|} p(\hat{y}_t^{(i)} | \hat{y}_{<t}^{(i)}, x)$$

    $$p_\text{softmax}(\hat{y}^{(i)}) = \frac{\exp(p(\hat{y}^{(i)}))}{\sum_{j=1}^k \exp(p(\hat{y}^{(j)}))}$$
    
    \begin{equation}
        \label{Eq:beam_entropy}
        \text{Conf}_{\text{Beam Entropy}} = -\sum_{i=1}^k p_\text{softmax}(\hat{y}^{(i)}) \text{ln} (p_\text{softmax}(\hat{y}^{(i)}))
    \end{equation}

    \begin{equation}
        \label{Eq:top_k_probs}
        \text{Conf}_{\text{Top K Probs}} = \sum_{i=1}^k p(\hat{y}^{(i)}) 
    \end{equation}
}

\begin{table*}[!ht]\centering
\resizebox{\linewidth}{!}{%
    \begin{tabular}{lrrrrrrrrrrrrrrrrrrr}\toprule
& &\multicolumn{2}{c}{Fil–EN} &\multicolumn{2}{c}{DE–EN} &\multicolumn{2}{c}{RU–EN} &\multicolumn{2}{c}{HotpotQA} &\multicolumn{2}{c}{SQUAD} &\multicolumn{2}{c}{Debate} &\multicolumn{2}{c}{Reddit} &\multicolumn{2}{c}{CNN} & \multicolumn{2}{c}{XSUM} \\\cmidrule{3-20}
& &Bt &FT5 &Bt &FT5 &Bt &FT5 &Bt &FT5 &Bt &FT5 &Bt &FT5 &Bt &FT5 &Bt &FT5 &Bt &FT5 \\
\midrule
\parbox[t]{4mm}{\multirow{2}{*}{\rotatebox[origin=c]{90}{Ours}}} & Tail & 0.649 & 0.380 & 0.648 & 0.190 & 0.779 & 0.506 & 0.255 & 0.451 & 0.494 & 0.582 & 0.518 & 0.354 & 0.601 & 0.300 & 0.100 & 0.031 & 0.131 & 0.212 \\
& Ratio & 0.546 & 0.200 & 0.653 & 0.209 & 0.768 & 0.491 & 0.249 & 0.360 & 0.505 & 0.565 & 0.496 & 0.293 & 0.596 & 0.304 & 0.103 & 0.055 & 0.082 & 0.196 \\
\cmidrule{2-20}
\parbox[t]{4mm}{\multirow{2}{*}{\rotatebox[origin=c]{90}{Base}}} & Entropy & 0.649 & 0.381 & 0.648 & 0.190 & 0.779 & 0.507 & 0.256 & 0.448 & 0.472 & 0.565 & 0.518 & 0.354 & 0.594 & 0.346 & 0.101 & 0.043 & 0.135 & 0.215 \\
& Sum Top K & 0.516 & 0.495 & 0.165 & 0.285 & 0.603 & 0.058 & 0.103 & 0.144 & 0.133 & 0.006 & 0.490 & 0.253 & 0.616 & 0.575 & 0.106 & 0.162 & 0.120 & 0.063 \\
\bottomrule
\end{tabular}
}
\caption{Spearman correlation (absolute value) between confidence score and evaluation metric/quality score (BLEU for translation, F1 for QA, RougeL for summarization); Bt: BART, FT5: Flan-T5}
\label{Table:Additional_Baselines}
\end{table*}

\section{Failure Case Examples}
\label{Appendix:Failure_Cases}

We provide examples of cases where there is miscalibration, either due to actual model miscalibration (Table \ref{Table:Miscalibrated_Model}), or due to issues with the evaluation strategy (Table \ref{Table:Miscalibrated_Label}).

\begin{table*}[]
\scriptsize
\centering
\begin{tabular}{p{15cm}}
\toprule
\textbf{Overconfident Model: Wrong Translation} \\
\midrule
\textbf{Source:} Translate English to Filipino: In the archipelagos and lakes you do not necessarily need a yacht \\
\midrule
\textbf{Prediction:} Ang mga archipelago at mga lupa ay hindi nangangailangan ng isang yacht. (\textit{Archipelagos and \textbf{\textcolor{red}{land}} do not need a \textbf{\textcolor{red}{yacht}}}) \\
\midrule
\textbf{Target:} Sa mga arkipelago at mga lawa ay hindi mo naman palaging kakailanganin ang yate. (\textit{In archipelagos and lakes, you do not always need a yacht.}) \\
\toprule
\textbf{Overconfident Model: Wrong Translation}\\
\midrule
\textbf{Source:} Translate English to Filipino: Scotturb Bus 403 travels regularly to Sintra, stopping at Cabo da Roca \\
\midrule
\textbf{Prediction:} Ang Scotturb Bus 403 ay nagsimula sa Sintra, na nagsimula sa Cabo da Roca. (\textit{The Scotturb Bus 403 \textbf{\textcolor{red}{starts}} from Sintra, and \textbf{\textcolor{red}{starts}} from Cabo de Roca})\\
\midrule
\textbf{Target:} Regular na bumibiyahe ang Scotturb Bus 403 patungong Sintra, tumitigil sa Cabo da Roca. (\textit{The Scotturb Bus 403 regularly travels to Sintra, stopping at Cabo da Roca}) \\
\bottomrule
\end{tabular}
\caption{Examples of outputs where the confidence scores themselves are miscalibrated, taken from the FLORES (Filipino) Dataset \citep{nllb-2022} \label{Table:Miscalibrated_Model}}
\end{table*}

\begin{table*}[]
\scriptsize
\centering
\begin{tabular}{p{15cm}}
\toprule
\textbf{Good Output Rated as Bad: Correct Gist, Different Style } \\
\midrule
\textbf{Source:} Manchester United winger Ashley Young celebrated his prominent role in his side's derby win by laughing at 'noisy neighbours' City. The 29-year-old, who scored United's first goal, and created two more for Marouane Fellaini and Chris Smalling, said Louis van Gaal's side were focused on quieting the Manchester City fans... 'Confidence has been sky high for a long time and we want to keep the momentum going now. ... I think it was over when it went to 3-1 \\
\midrule
\textbf{Prediction:} Ashley Young celebrated his prominent role in his side's derby win by laughing at 'noisy neighbours' City fans. The 29-year-old, who scored United's first goal, and created two more for Marouane Fellaini and Chris Smalling, said Louis van Gaal's side were focused on quieting the Manchester City fans. \\
\midrule
\textbf{Target:} Ashley Young scored Manchester United equaliser in 4-2 win. 'We quietened them down straight away' claims Young, as he praises fans. Young claims the game was over once United went 3-1 ahead. England winger says confidence at Old Trafford is now 'sky high'. CLICK HERE to read Ian Ladyman's match report from Old Trafford. READ: Man Utd runaway league leaders in table vs the current top seven\\
\toprule
\textbf{Good Output Rated as Bad: Noisy Label}\\
\midrule
\textbf{Source:} My first trip to New York was with my aunt when I was about 18 and I had been appearing in Coronation Street for about 12 months... we went to see the musical Chicago on Broadway and did lots and lots of shopping... one of the first things my boyfriend Sylvain Longchambon and I did on my latest trip was to visit the site of the World Trade Center to see the monument to those who died... Coronation Street star Samia Ghadie and boyfriend Sylvain Longchambon take a selfie in New York... The JW Marriott Essex House as seen from Central Park, of which it boasts panoramic views... On the second night of our break, we went on the Bateaux New York cruise, which offers an unforgettable three-hour trip along the Hudson River... The couple, who met as partners on Dancing on Ice, enjoyed a Bateaux New York Cruise of the Hudson...We were wondering who was going to emerge when out stepped Lady Gaga. It was the ultimate holiday treat - it was just a shame I couldn't arrange a selfie with her. That would have been the icing on the cake. American Sky (0843 636 2001) offers a three-night holiday to New York from £899pp. This includes accommodation in the JW Marriott Essex House New York on a room-only basis and return flights with British Airways from Heathrow. Visit nycgo com for more information \\
\midrule
\textbf{Prediction:} My first trip to New York was with my aunt when I was about 18 and I had been appearing in Coronation Street for about 12 months. We had a lot of fun together - we went to see the musical Chicago on Broadway and did lots and lots of shopping. We returned to the city that never sleeps in January 2001, eight months before the terrible events of 9/11, and one of the first things my boyfriend Sylvain Longchambon and I did on my latest trip was to visit the site of the World Trade Center to see the monument to those who died.\\
\midrule
\textbf{Target:} Samia Ghadie and boyfriend Sylvain Longchambon spent four days in NYC. While there the Dancing on Ice stars stayed at the JW Marriott Essex House. Their suite on the 31st floor offered up panoramic views of Central Park. The couple enjoyed a three-hour Bateaux New York cruise on the Hudson. The confirm Burger Joint in Le Parker Meridien Hotel lives up to the hype.\\
\toprule
\textbf{Bad Output Rated as Good: Repeated Phrase}\\
\midrule
\textbf{Source:} SOURCE:
An exotic animal owner from Ohio is campaigning for his bear and big cats to be returned after they were seized by the state in January. In total, six tigers, a bear, a lion, a cougar, a black leopard and a liger (part lion, part tiger) were taken from Kenny Hetrick's Stony Ridge farm after it was found he did not have the correct permit and cages were 'unsafe'. Authorities have also started cracking down on the owners of wild creatures following an incident in 2011 where a man in eastern Ohio released 56 exotic animals - including lions and tigers - then killed himself. Apparently Hetrick has been heartbroken by the clampdown on his farm and the 72-year-old widower is now fighting to overturn the seizure, ... 'He's lost without them,' said Josh Large, who lives four houses away... \\
\midrule
\textbf{Prediction:} \textcolor{blue}{In total, six tigers, a bear, a lion, a cougar, a black leopard and a liger were taken from Kenny Hetrick's Stony Ridge farm.
In total, six tigers, a bear, a lion, a cougar, a black leopard and a liger were taken from Kenny Hetrick's Stony Ridge farm} after it was found he did not have the correct permit and cages were 'unsafe'. Authorities have also started cracking down on the owners of  wild creatures following an incident in 2011 where a man in eastern Ohio released 56 exotic animals - including lions and  tigers - then killed himself.
Hetrick has been heartbroken by the clampdown on his farm and the 72-year-old widower is now fighting to overturn the seizure, backed by neighbors who insist his menagerie doesn't pose a threat. \\
\midrule
\textbf{Target:} In total, six tigers, a bear, a lion, a cougar, a black leopard and a liger (part lion, part tiger) were taken from Kenny Hetrick's Stony Ridge farm .
State officials found he didn't have the right permit and cages were 'unsafe'
But now the 72-year-old is fighting to overturn the seizure, backed by neighbors who insist his menagerie doesn't pose a threat .
'He's lost without them,' said Josh Large, who lives four houses away .\\

\bottomrule
\end{tabular}
\caption{Examples of outputs where the outputs are rated incorrectly based on the metric, taken from the CNN-DailyMail Dataset \citep{see-etal-2017-get} \label{Table:Miscalibrated_Label}}
\end{table*}

\end{document}